\documentclass{article}

\usepackage{microtype}
\usepackage{graphicx}
\usepackage{booktabs} 
\usepackage{subcaption}

\usepackage{hyperref}
\usepackage{amsmath,amsfonts,amssymb,amsthm,mathrsfs}

\usepackage{pgfplots,tikz}
\newlength{\figurewidth}
\newlength{\figureheight}
\pgfplotsset{/pgf/number format/1000 sep={}}

\usepackage{bm}
\newcommand{\mathbold}[1]{\bm{#1}}
\newcommand{\mbf}[1]{\mathbf{#1}}

\newcommand{\T}{\top}

\newcommand{\valpha}[0]{\mathbold{\alpha}}
\newcommand{\vbeta}[0]{\mathbold{\beta}}
\newcommand{\vgamma}[0]{\mathbold{\gamma}}

\newcommand{\vrho}[0]{\mathbold{\rho}}

\newcommand{\vmu}[0]{\mathbold{\mu}}

\newcommand{\vtheta}[0]{\mathbold{\theta}}

\newcommand{\vb}{\mbf{b}}

\newcommand{\vf}{\mbf{f}}

\newcommand{\vk}{\mbf{k}}

\newcommand{\vm}{\mbf{m}}

\newcommand{\vr}{\mbf{r}}

\newcommand{\vu}{\mbf{u}}

\newcommand{\vw}{\mbf{w}}

\newcommand{\vy}{\mbf{y}}
\newcommand{\vz}{\mbf{z}}
\newcommand{\MA}{\mbf{A}}
\newcommand{\MB}{\mbf{B}}

\newcommand{\MF}{\mbf{F}}
\newcommand{\MG}{\mbf{G}}
\newcommand{\MH}{\mbf{H}}
\newcommand{\MI}{\mbf{I}}

\newcommand{\MK}{\mbf{K}}
\newcommand{\ML}{\mbf{L}}

\newcommand{\MP}{\mbf{P}}
\newcommand{\MQ}{\mbf{Q}}
\newcommand{\MR}{\mbf{R}}

\newcommand{\MT}{\mbf{T}}
\newcommand{\MV}{\mbf{V}}
\newcommand{\MW}{\mbf{W}}
\newcommand{\MX}{\mbf{X}}

\definecolor{cgray}{gray}{0.4}
\newcommand{\comm}[1]{\hfill\textcolor{cgray}{\# #1}}

\usepackage[accepted]{icml2018-arxiv}

\icmltitlerunning{Non-Gaussian State Space GPs}

\begin{document}

\twocolumn[
\icmltitle{State Space Gaussian Processes with Non-Gaussian Likelihood}

\icmlsetsymbol{equal}{*}

\begin{icmlauthorlist}
\icmlauthor{Hannes Nickisch}{philips}
\icmlauthor{Arno Solin}{aalto}
\icmlauthor{Alexander Grigorievskiy}{aalto,silo}
\end{icmlauthorlist}

\icmlaffiliation{philips}{Digital Imaging, Philips Research, Hamburg, Germany}
\icmlaffiliation{aalto}{Department of Computer Science, Aalto University, Espoo, Finland}
\icmlaffiliation{silo}{Silo.AI, Helsinki, Finland}

\icmlcorrespondingauthor{Hannes Nickisch}{hannes@nickisch.org}
\icmlcorrespondingauthor{Arno Solin}{arno.solin@aalto.fi}

\icmlkeywords{Gaussian Processes, Kalman Filter, State Space, Approximate Inference, Time Series}

\vskip 0.3in
]

\printAffiliationsAndNotice{ \textit{Proceedings of the $\mathit{35}^{th}$ International Conference on Machine Learning},
Stockholm, Sweden, PMLR 80, 2018. Copyright 2018 by the author(s).}  

\begin{abstract}
We provide a comprehensive overview and tooling for GP modeling with non-Gaussian likelihoods using state space methods. The state space formulation allows to solve one-dimensional GP models in $\mathcal{O}(n)$ time and memory complexity. While existing literature has focused on the connection between GP regression and state space methods, the computational primitives allowing for inference using general likelihoods in combination with the Laplace approximation (LA), variational Bayes (VB), and assumed density filtering (ADF, a.k.a.\ single-sweep expectation propagation, EP) schemes has been largely overlooked. We present means of combining the efficient $\mathcal{O}(n)$ state space methodology with existing inference methods. We extend existing methods, and provide unifying code implementing all approaches.
\end{abstract}

\section{Introduction}
\label{sec:intro}
Gaussian processes (GPs) \cite{rasmussen06} form a versatile class of probabilistic machine learning models with applications in regression, classification as well as robust and ordinal regression. In practice, there are computational challenges arising from {\em (i)} non-conjugate (non-Gaussian) likelihoods and {\em (ii)} large datasets.

The former {\em (i)} can be addressed by approximating the non-Gaussian posterior by an effective Gaussian giving rise to a number of algorithms such as the Laplace approximation \citep[LA,][]{williams98laplace}, variational Bayes \citep[VB,][]{gibbs00vb}, direct Kullback--Leibler (KL) divergence minimization \cite{opper09kl} and expectation propagation \citep[EP,][]{minka01} with different tradeoffs in terms of accuracy and required computations \cite{kuss05assessing, nickisch08approximations, naish-guzman2008}.
The latter {\em (ii)} can be addressed by approximate covariance computations using sparse inducing point methods \cite{quinonero05unifying} based on variational free energy  \citep[VFE,][]{titsias2009vfe}, fully independent training conditionals \citep[FITC,][]{snelson06FITC}, hybrids \cite{bui16spep}, or stochastic approximations \cite{Hensman+Fusi+Lawrence:2013,krauth2017} applicable to any data dimension $D$. A second class of covariance interpolation methods, KISS-GP \cite{wilson15kiss-gp,wilson15msgp}, are based on grids of inducing points.  For $1<D<5$, product covariance, and rectilinear grids, the covariance matrix has Kronecker structure. For $D=1$, stationary covariance, and a regular grid, the covariance matrix has Toeplitz structure (a special case of block-Toeplitz with Toeplitz blocks (BTTB) obtained for $1<D<5$), which can be exploited for fast matrix-vector multiplications (MVMs).
A third covariance approximation methodology is based on basis function expansions such as sparse spectrum GPs \cite{Lazaro-Gredilla+Quinonero-Candela+Rasmussen+Figueiras-Vidal:2010}, variational Fourier features \cite{hensman2016variational}, or Hilbert space GPs  \cite{Solin+Sarkka:2014b} for stationary covariance functions. Higher input dimensions $D>4$ either tend to get computationally heavy or prone to overfitting.

In time-series data, with $D=1$, the data sets tend to become long (or unbounded) when observations accumulate over time. For these time-series models, leveraging sequential \emph{state space} methods from signal processing makes it possible to solve GP inference problems in {\em linear time complexity} $\mathcal{O}(n)$ if the underlying GP has Markovian structure \citep{Reece+Roberts:2010, Hartikainen+Sarkka:2010}. This reformulation is {\em exact} for Markovian covariance functions \citep[see, e.g.,][]{Solin:2016} such as the exponential, half-integer Mat\'ern, noise, constant, linear, polynomial, Wiener, etc.\ (and their sums and products). Covariance functions such as the squared exponential \citep{Hartikainen+Sarkka:2010}, rational quadratic \citep{Solin+Sarkka:2014-MLSP}, and periodic \citep{Solin+Sarkka:2014} can be approximated by their Markovian counterparts. \citet{grigorievskiy16gp_ss,grigorievskiy17sparseGPprecision} bridge the state space connection further by leveraging sparse matrices (SpInGP) in connection with the Markovian state space models. Another issue is that if time gaps between data points are very uneven then the computational power is spend on computing required matrix exponentials. This still makes the method slow for the large datasets with uneven sampling despite the linear computational complexity of inference. This shows as a large cost per time step (the `hidden' constant in the big-O notation) due to evaluating matrix exponentials.

The previous literature has focused on rewriting the GP in terms of a state space model (focusing on challenge {\em (i)}). Addressing challenge {\em (ii)}, non-Gaussian likelihoods have been touched upon by \citet{Solin+Sarkka:2014-MLSP} (inner-loop Laplace approximation) and \citet{Hartikainen+Riihimaki+Sarkka:2011} in a spatio-temporal log Gaussian Cox process (using EP combined with local extended Kalman filtering updates). However, deriving approximate inference schemes is in the state space regime is complicated and requires hand-crafting for each likelihood.

Related work also includes Kalman filtering for optimization in parametric models \cite{aravkin13sparseKalman, aravkin14optimizationKalman}, and non-linear GP priors in system identification models \citep[a.k.a.\ `GP state space' models, see, e.g.,][]{frigola14variational_gp_ss}.

This paper advances the state-of-the-art in two ways:
\begin{itemize}
  \item We present a unifying framework for solving computational primitives for non-Gaussian inference schemes in the state space setting, thus directly enabling inference to be done through LA, VB, KL, and ADF/EP.
  \item We present a novel way for solving the continuous-time state space model through interpolation of the matrix exponential, which further speeds up the linear time-complexity by addressing the large-constant problem.
\end{itemize}
Code for the paper is available as part of the GPML toolbox version 4.2 \cite{rasmussen10gpml}.

\section{Methods}
\label{sec:methods}
We introduce the GP framework in Sec.~\ref{sub:gp}, then name four computational primitives that can be used to operate approximate inference schemes beyond the exact Gaussian case in Sec.~\ref{sub:primitives}. 
The state space representation of GPs is introduced in \ref{sub:gpss} along with the Kalman filtering and smoothing algorithms, Algs.~\ref{alg:kalman}+\ref{alg:rts}.
Then, we will show how these primitives including prediction can be implemented for GPs using the state space representation in Sec.~\ref{sub:ss_primitives}.
Further, we detail how they can be used to operate inference for Laplace approximation (LA) in Sec.~\ref{sub:la}, variational Bayes (VB) in Sec.~\ref{sub:vb}, assumed density filtering (ADF) a.k.a.\ single sweep expectation propagation (EP) in Sec.~\ref{sub:adf} and Kullback--Leibler (KL) minimization in Sec.~\ref{sub:kl}. For the first three algorithms, we are also able to perform full-fledged gradient-based hyperparameter learning.

\subsection{Gaussian process training and prediction}
\label{sub:gp}
The models we are interested, in take the following standard form of having a latent Gaussian process prior and a measurement (likelihood) model:
\begin{equation*}
  f(t)\sim\mathrm{GP}(m(t),k(t,t')),\quad\vy|\vf\sim\prod_{i=1}^{n}\mathbb{P}(y_{i}|f(t_{i})),
\end{equation*}
where the likelihood factorizes over the observations. This family of models covers many types of modeling problems including (robust or ordinal) regression and classification.

We denote the data as a set of scalar input--output pairs $\mathcal{D} = \{(t_i,y_i)\}_{i=1}^n$. We are interested in models following \citet{rasmussen10gpml} that -- starting from the Gaussian prior $\vf = \mathrm{N}(\vf | \vm, \MK)$ given by the GP -- admit an approximate posterior of the form
\begin{equation}
  \mathbb{Q}(\vf | \mathcal{D}) = \mathcal{N}\left(\vf|\vm+\MK\valpha,(\MK^{-1}+\MW)^{-1}\right),
\end{equation}
where $m_i = m(t_i)$ and $K_{i,j} = k(t_i,t_j)$ are the prior mean and covariance. The vector $\valpha$ and the (likelihood precision) matrix $\MW=\text{diag}(\vw)$ form the set of $2n$ parameters.
Elements of $\vw$ are non negative for log-concave likelihoods.
Equivalently, we can use the natural parameters $(\vb,\MW)$ of the effective likelihood, where $\vb=\MW\MK\valpha+\valpha$ in general and for Gaussian likelihood $\vb=\MW(\vy-\vm)$ in particular.

Given these parameters, the predictive distribution for an unseen test input $t_*$ is obtained by integrating the Gaussian latent marginal distribution $\mathcal{N}(f_*|\mu_{f,*},\sigma^2_{f,*})$
\begin{equation}
  \mu_{f,*} \negmedspace=\negmedspace \vm_* \negmedspace+\negmedspace \vk_*^\T \valpha;\:
	\sigma^2_{f,*}\negmedspace=\negmedspace k_{**}\negmedspace -
	                 \vk_*^\T \left( \MK \negmedspace+\negmedspace \MW^{-1} \right)^{-1} \vk_*
\end{equation}
against the likelihood $\mathbb{P}(y_*|f_*)$ to obtain
\begin{equation}
  \mathbb{P}(y_*) = \int \mathbb{P}(y_*|f_*) \, \mathcal{N}(f_*|\mu_{f,*},\sigma^2_{f,*}) \,\mathrm{d}f_*
\end{equation}
the predictive distribution whose first two moments can be used to make a statement about the unknown $y_*$.

The model may have hyperparameters $\vtheta=[a,d,\sigma_f,\ell,\sigma_n]$ of the mean e.g.\ $m(t)=at+d$, the covariance e.g.\ $k(t,t')=\sigma_f^2 \exp(-(t-t')^2/(2\ell^2))$ and the likelihood e.g.\ $\mathbb{P}(y_i|f_i)=\mathcal{N}(f_i|y_i,\sigma_n^2)$ which can be fit by maximizing the (log) marginal likelihood of the model
\begin{equation}
  \log Z(\vtheta) = \log \int \mathcal{N}\left(\vf|\vm,\MK\right) \prod_i \mathbb{P}(y_i|f_i) \,\mathrm{d} \vf,
\end{equation}
which is an intractable integral in the non-Gaussian case but can be approximated or bounded in various ways.

\begin{algorithm}[tb]
   \caption{Predictions and log marginal likelihood $\log Z$ for Gaussian process regression (Alg.~2.1 in \citet{rasmussen06}). Complexity is $\mathcal{O}(n^3)$ for the Cholesky decomposition, and $\mathcal{O}(n^2)$ for solving triangular systems.}
   \label{alg:gpreg}
\begin{algorithmic}
   \STATE {\bfseries Input:} $\{t_i\},\:\{y_i\}$ \comm{training inputs and targets}\\                             $k,\:\sigma_n^2,\:t_*$ \comm{covariance, noise variance, test input}
   \STATE $\ML  \leftarrow  \mathrm{Cholesky}(\MK + \sigma_\mathrm{n}^2 \, \MI);\:\:\valpha  \leftarrow  \ML^{-\top}  (\ML^{-1} (\vy-\vm))$
   \STATE $\log Z  \leftarrow  -\frac{1}{2}(\vy-\vm)^\top \valpha - \sum_i \log L_{i,i} - \frac{n}{2} \log 2\pi$
   \STATE $\mu_{f,*} \leftarrow \vm_* + \vk_*^\top \valpha;\:\:\sigma^2_{f,*} \leftarrow  k_{**} -  \left\Vert \ML \backslash \vk_* \right\Vert^2_2$
   \STATE {\bfseries Return:} $\mu_{f,*},\:\sigma^2_{f,*},\:\log Z$ \comm{mean, variance, evidence}
\end{algorithmic}
\end{algorithm}

A prominent instance of this setting is plain GP regression (see Alg.~\ref{alg:gpreg}), where the computation is dominated by the $\mathcal{O}(n^3)$ log-determinant computation and the linear system for $\valpha$. To overcome the challenges arising from non-conjugacy and large dataset size $n$, we define a set of generic computations and replace their dense matrix implementation (see Alg.~\ref{alg:gpreg}) with state space algorithms.

\subsection{Gaussian process computational primitives}
\label{sub:primitives}
The following computational primitives allow to cast the covariance approximation in more generic terms:
\begin{enumerate}
  \item Linear system with ``regularized'' covariance: $\text{solve}_{\MK}(\MW,\vr):=(\MK+\MW^{-1})^{-1}\vr$.
  \item Matrix-vector multiplications: $\text{mvm}_{\MK}(\vr):=\MK\vr$.\\
	      For learning we also need $\frac{\text{mvm}_{\MK}(\vr)}{\partial\vtheta}$.
  \item Log-determinants: $\text{ld}_{\MK}(\MW):=\log|\MB|$ with symmetric and 
	      well-conditioned $\MB=\mathbf{I}+\MW^{\frac{1}{2}}\MK\MW^{\frac{1}{2}}$. \\
        For learning, we need derivatives:
				$\frac{\partial\text{ld}_{\MK}(\MW)}{\partial\vtheta}$,
        $\frac{\partial\text{ld}_{\MK}(\MW)}{\partial\MW}$.
	\item Predictions need latent mean $\mathbb{E}[f_*]$ and variance $\mathbb{V}[f_*]$.
\end{enumerate}
Using these primitives, GP regression can be compactly written as $\MW=\MI/\sigma^2_n$, $\valpha=\text{solve}_{\MK}(\MW,\vy-\vm)$, and 
\begin{multline}
  \log Z_\mathrm{GPR} = \\
  -\frac{1}{2}\left[\valpha^{\top}\text{mvm}_{\MK}(\valpha)
                    +\text{ld}_{\MK}(\MW)
                    +n\log(2\pi\sigma_{n}^{2})\right].
\end{multline}

Approximate inference (LA, VB, KL, ADF/EP) -- in case of non-Gaussian likelihoods -- requires these primitives as necessary building blocks. Depending on the covariance approximation method e.g.\ exact, sparse, grid-based, or state space, the four primitives differ in their implementation and computational complexity.

\begin{algorithm}[tb]
   \caption{Kalman (forward) filtering. For ADF, $(\MW,\vb)$ are not required as inputs.
	          Note, $\vb = \MW \vr$.}
   \label{alg:kalman}
\begin{algorithmic}
   \STATE {\bfseries Input:} $\{t_i\}$ , $\vy$                    \comm{training inputs and targets} \\
	               \quad \qquad $\{\MA_i\}$, $\{\MQ_i\}$, $\MH$, $\MP_0$      \comm{state space model} \\
								 \quad \qquad $\MW$, $\vb$             \comm{likelihood eff.\ precision and location}
   \FOR{$i=1$ {\bfseries to} $n$}
     \IF{$i==1$}
		   \STATE $\vm_i \leftarrow \bm{0};\:\: \MP_i \leftarrow \MP_0$                        \comm{init}
		 \ELSE
       \STATE $\vm_i \leftarrow \MA_{i} \vm_{i-1};\: 
			         \MP_i \leftarrow \MA_{i} \MP_{i-1} \MA_{i}^{\top}\negmedspace+\negmedspace\MQ_{i}$ \comm{predict}
     \ENDIF
		 \IF{has label $y_i$}
		   \STATE $\mu_f \leftarrow \MH \vm_i;\:\: \vu \leftarrow \MP_i \MH^\T;\:\: 
			         \sigma^2_f \leftarrow \MH \vu$                                              \comm{latent}
			 \IF{ADF (assumed density filtering)}
			    \STATE set $(b_i,W_{ii})$ to match moments of $\mathbb{P}(y_i|f_i)$ and 
					       $\exp(b_{i}f_i-W_{ii}f_i^{2}/2)$ w.r.t. latent $\mathcal{N}(f_i|\mu_{f},\sigma_{f}^{2})$
			 \ENDIF
			 \STATE $z_i \leftarrow W_{ii} \sigma^2_f + 1;\:\:c_i \leftarrow W_{ii} \mu_f - b_i$

			 \STATE $\vk_i \leftarrow W_{ii} \vu/z_i;\:\: \MP_i \leftarrow \MP_i - \vk_i \vu^\T$  \comm{variance}
			 \STATE $\gamma_i \leftarrow - c_i/z_i;\:\:\vm_i \leftarrow \vm_i + \gamma_i \vu$     \comm{mean}
		 \ENDIF
   \ENDFOR
	 \STATE $\text{ld}_{\MK}(\MW) \leftarrow \sum_i \log z_{i}$
\end{algorithmic}
\end{algorithm}

\begin{algorithm}[tb]
   \caption{Rauch--Tung--Striebel (backward) smoothing.}
   \label{alg:rts}
\begin{algorithmic}
   \STATE {\bfseries Input:} $\{\vm_i\}$, $\{\MP_i\}$ \comm{Kalman filter output}\\
	              \quad \qquad $\{\MA_i\}$, $\{\MQ_i\}$ \comm{state space model}
   \FOR{$i=n$ {\bfseries down to} $2$}
	   \STATE $\vm \leftarrow \MA_i \vm_{i-1} ;\:\: 
		         \MP \leftarrow \MA_i \MP_{i-1} \MA_i^\T + \MQ_i$                        \comm{predict}
		 \STATE $\MG_i \leftarrow \MP_{i-1} \MA_i^\T \MP^{-1};\:\: 
		         \Delta\vm_{i-1}\leftarrow \MG_i (\vm_{i} - \vm)$
     \STATE $\MP_{i-1} \leftarrow \MP_{i-1} + \MG_i (\MP_{i} - \MP) \MG_i^\T$        \comm{variance}
     \STATE $\vm_{i-1} \leftarrow \vm_{i-1} + \Delta\vm_{i-1}$                       \comm{mean}
		 \STATE $\rho_{i-1} \leftarrow \MH \Delta\vm_{i-1}$                              \comm{posterior}
   \ENDFOR
	 \STATE $\text{solve}_{\MK}(\MW,\vr) = \valpha \leftarrow \vgamma - \MW \vrho$     \comm{posterior}      
\end{algorithmic}
\end{algorithm}

\subsection{State space form of Gaussian processes}
\label{sub:gpss}
GP models with covariance functions with a Markovian structure can be transformed into equivalent state space models. The following exposition is based on \citet[Ch.~3]{Solin:2016}, which also covers how to derive the equivalent exact models for sum, product, linear, noise, constant, Mat\'ern (half-integer), Ornstein--Uhlenbeck, and Wiener covariance functions. Other common covariance functions can be approximated by their Markovian counterparts, including squared exponential, rational quadratic, and periodic covariance functions.

A state space model describes the evolution of a dynamical system at different time instances $t_{i},\:i=1,2,\ldots$ by
\begin{equation}
  \vf_{i} \sim \mathbb{P}(\vf_{i} | \vf_{i-1}), \quad y_{i}\sim\mathbb{P}(y_{i} | \vf_{i}),
\end{equation}
where $\vf_{i}:=\vf(t_{i})\in\mathbb{R}^{d}$ and $\vf_{0}\sim\mathbb{P}(\vf_{0})$ with $\vf_{i}$ being the latent (hidden/unobserved) variable and $y_{i}$ being the observed variable. In continuous time, a simple dynamical system able to represent many covariance functions is given by the following linear time-invariant stochastic differential equation:
\begin{equation}
  \dot{\vf}(t)=\MF\,\vf(t)+\ML\,\vw(t),  \quad y_{i}=\MH\,\vf(t_{i})+\epsilon_{i},
\end{equation}
where $\vw(t)$ is an $s$-dimensional white noise process, the measurement noise $\epsilon_{i}\sim\mathcal{N}(0,\sigma_{n}^{2})$ is Gaussian, and $\MF\in\mathbb{R}^{d\times d}$, $\ML\in\mathbb{R}^{d\times s}$, $\MH\in\mathbb{R}^{1\times d}$ are the feedback, noise effect, and measurement matrices, respectively. The initial state is distributed according to $\vf_{0}\sim\mathcal{N}(\mathbf{0},\MP_{0})$. 

The latent GP is recovered by $f(t)=\MH\vf(t)$ and $\vw(t)\in\mathbb{R}^{s}$ is a multivariate white noise process with spectral density matrix $\MQ_{c}\in\mathbb{R}^{s\times s}$. For discrete values, this translates into
\begin{equation}
  \vf_{i}\sim\mathcal{N}(\MA_{i-1}\vf_{i-1},\MQ_{i-1}), \quad 
   y_{i}\sim\mathbb{P}(y_{i} | \MH\,\vf_{i}),
\end{equation}
with $\vf_{0}\sim\mathcal{N}(\mathbf{0},\MP_{0})$. The discrete-time matrices are
\begin{align}
  \MA_{i} &= \MA[\Delta t_{i}] = e^{\Delta t_{i}\MF}, \\
  \MQ_{i} &= \int_{0}^{\Delta t_{i}}e^{(\Delta t_{k}-\tau)\MF}\ML\,\MQ_{c}\,\ML^{\top}e^{(\Delta t_{i}-\tau)\MF^{\top}}\mathrm{d}\tau,
\end{align}
where $\Delta t_{i}=t_{i+1}-t_{i}\ge0$.

For stationary covariances $k(t,t')=k(t-t')$, the stationary state is distributed by $\vf_{\infty}\sim\mathcal{N}(\mathbf{0},\MP_{\infty})$ and the stationary covariance can be found by solving the Lyapunov equation
\begin{equation}
  \dot{\MP}_{\infty}=\MF\,\MP_{\infty}+\MP_{\infty}\,\MF^{\top}+\ML\,\MQ_{c}\,\ML^{\top}=\mathbf{0},
\end{equation}
which leads to the identity $\MQ_{i}=\MP_{\infty}-\MA_{i}\,\MP_{\infty}\,\MA_{i}^{\top}$.

\subsection{Fast computation of $\MA_i$ and $\MQ_i$ by interpolation}
\label{sub:fast_interp}
In practice, the evaluation of the $n$ discrete-time transition matrices $\MA_i=e^{\Delta t_i \MF}$ and the noise covariance matrices $\MQ_i=\MP_{\infty}-\MA_{i}\MP_{\infty}\MA_i^{\top}$  (in the stationary case) for different values of $\Delta t_i$ is a computational challenge. When the distribution of $\Delta t_i$ in the dataset is narrow then computed matrices can be reused. However, when the distribution is wide, then computing $\MA_i$ and $\MQ_i$ consumes roughly 50\% of the time on average if done na\"{i}vely.

Since the matrix exponential $\psi:s\mapsto e^{s \MX}$ is smooth, its evaluation can be accurately approximated by convolution interpolation \cite{keys1981} as done for the covariance functions in the KISS-GP framework \cite{wilson15kiss-gp,wilson15msgp}.
The idea is to evaluate the function on a set of equispaced discrete locations $s_{1},s_{2},..,s_{K}$, where $s_{j}=s_{0}+j\cdot\Delta s$ and interpolate $\MA=e^{s\MX}$ from the closest precomputed $\MA_{j}=e^{s_{j}\MX}$ i.e.\ use the 4 point approximation $\MA\approx c_{1}\MA_{j-1}+c_{2}\MA_{j}+c_{3}\MA_{j+1}+c_{4}\MA_{j+2}$. The grid resolution $\Delta s$ governs approximation accuracy.

The same interpolation can be done for the noise covariance matrices $\MQ_i$. Finally, the number of matrix exponential evaluations can be reduced from $n$ to $K$, which -- for large datasets -- is practically negligible. The accuracy of the interpolation depends on the underlying grid spacing $\Delta s$. In practice, we use an equispaced grid covering range $[\min_i \Delta t_i, \max_i \Delta t_i]$, but hybrid strategies, where the bulk of the mass of the $\Delta t_i$ is covered by the grid and outliers are evaluated exactly, are -- of course -- possible. Very diverse sets of $\Delta t_i$ with vastly different values, could benefit from a clustering with an individual grid per cluster.

\subsection{State space computational primitives}
\label{sub:ss_primitives}

In the following, we will detail how the SpInGP viewpoint of \citet{grigorievskiy17sparseGPprecision} can be used to implement the computational primitives of Sec.~\ref{sub:primitives} with linear complexity in the number of inputs $n$. The covariance matrix of the latent GP $f(t)$ evaluated at the training inputs $t_{1},\ldots,t_{n}$ is denoted $\MK\in\mathbb{R}^{n\times n}$ and the (joint) covariance of the dynamical system state vectors $[\MF_{0};\MF_{1};..;\MF_{n}]$ is denoted by $\mathcal{K}\in\mathbb{R}^{(n+1)d\times(n+1)d}$. Defining the sparse matrix $\MG^{n\times(n+1)d}=[\mathbf{0}_{n\times d},\mathbf{I}_{n}\otimes\mathbf{H}]$, we obtain $\MK=\MG\mathcal{K}\MG^{\top}$. Further, define the symmetric block diagonal matrix
\[
\MQ=\left[\begin{array}{cccc}
\mathbf{P}_{0} & \mathbf{0} & \ldots & \mathbf{0}\\
\mathbf{0} & \MQ_{1} & \ldots & \mathbf{0}\\
\vdots & \vdots & \ddots & \vdots\\
\mathbf{0} & \mathbf{0} & \ldots & \MQ_{n}
\end{array}\right] \in \mathbb{R}^{(n+1)d\times(n+1)d}
\]
and $(n+1)d\times(n+1)d$ matrix $\MT=\MA^{-1}=$
\[
\left[\begin{array}{ccccc}
\mathbf{I} & \mathbf{0} & \mathbf{0} & \ldots & \mathbf{0}\\
-\MA[\Delta t_{1}] & \mathbf{I} & \mathbf{0} & \ldots & \mathbf{0}\\
\mathbf{0} & -\MA[\Delta t_{2}] & \mathbf{I} & \ldots & \mathbf{0}\\
\vdots & \vdots & \vdots & \ddots & \vdots\\
\mathbf{0} & \mathbf{0} & -\MA[\Delta t_{n}] & \ldots & \mathbf{I}
\end{array}\right]
\]
of block tridiagonal (BTD) structure allowing to write 
\[
\mathcal{K}^{-1}= \MT^{\top}\MQ^{-1}\MT ,\:\text{and}\:\mathcal{K}=\MA\MQ\MA^{\top},
\]
where it becomes obvious that $\mathcal{K}^{-1}$ is a symmetric BTD;
which is in essence the structure exploited in the SpInGP framework
by \citet{grigorievskiy17sparseGPprecision}.

\subsubsection{Linear systems}
\label{subsub:linsolve}
Using the the matrix inversion lemma, we can rewrite $\left(\MK+\MW^{-1}\right)^{-1}$
as
\begin{align*}
 & =\MW-\MW\MG\left(\mathcal{K}^{-1}+\MG^{\top}\MW^{-1}\MG\right)^{-1}\MG^{\top}\MW\\
 & =\MW-\MW\MG\MR^{-1}\MG^{\top}\MW,\:\MR= \MT^{\top}\MQ^{-1} \MT +\MG^{\top}\MW\MG.
\end{align*}
This reveals that we have to solve a system with a symmetric BTD system matrix $\MR$, where $\MG^{\top}\MW\MG=\text{diag}([0;\MW])\otimes(\mathbf{H}^{\top}\mathbf{H})$. The only (numerical) problem could be the large condition of any of the constituent matrices of $\MQ$ as it would render the multiplication with $\mathcal{K}^{-1}$ a numerical endeavour. Adding a small ridge $\alpha^{2}$ to the individual constituents of $\MQ$ i.e.\ use $\tilde{\MQ}_{i}=\MQ_{i}+\alpha^{2}\mathbf{I}$ instead of $\MQ_{i}$ is a practical remedy. Finally, we have
\[
\text{solve}_{\MK}(\MW,\MR)=\MW\MR-\MW\MG\MR^{-1}\ensuremath{\MG^{\top}\MW\MR}.
\]

\subsubsection{Matrix-vector multiplications}
\label{subsub:mvms}
Using the identity $\mathcal{K}=\MA\MQ\MA^{\top}$
from \citet{grigorievskiy17sparseGPprecision} and $\MK=\MG\mathcal{K}\MG^{\top}$,
we can write
\[
\text{mvm}_{\MK}(\MR)=\MG \MT^{-1} \MQ \MT^{-\top}\MG^{\top}\MR
\]
where all constituents allow for fast matrix-vector multiplications.
The matrix $\MG$ is sparse, the matrix $\MQ$ is block
diagonal and the linear system with $\MT$ is of BTD type.
Hence, overall runtime is $\mathcal{O}(nd^{2})$.
For the derivatives $\frac{\text{mvm}_{\MK}(\vr)}{\partial\theta_i}$, we proceed component-wise using
$\frac{\MQ}{\partial\theta_i}$ and
$\frac{\MT^{-1}}{\partial\theta_i}=-\MT^{-1} \frac{\MT}{\partial\theta_i} \MT^{-1}$.
The derivative $\text{d}\exp(\MX)$ of the matrix exponential $\exp(\MX)$ is obtained via a method 
by \citet[Eqs. 10\&11]{najfeld95expm_deriv} using a matrix exponential of twice the size
\[
\exp\left(\left[\begin{array}{cc}
\MX & \mathbf{0}\\
\text{d}\MX & \MX
\end{array}\right]\right)=\left[\begin{array}{cc}
\exp(\MX) & \mathbf{0}\\
\text{d}\exp(\MX) & \exp(\MX)
\end{array}\right].
\]

\subsubsection{Log determinants}
\label{subsub:logdet}
The Kalman filter (Alg.~\ref{alg:kalman}) can be used to compute the
log determinant $\text{ld}_{\MK}(\MW)=\sum_{i}\log z_{i}$
in $\mathcal{O}(nd^{3})$.

There are two kinds of derivatives of the log determinant required for learning (see Sec.~\ref{sub:primitives}). First, the hyperparameter derivatives 
$\frac{\partial\text{ld}_{\MK}(\MW)}{\partial\vtheta}$
are computed component-wise using a differential version of the Kalman filter (Alg.~\ref{alg:kalman}) as described in \citet[Appendix]{sarkka13Bayes_smooth_filter}, the matrix exponential derivative algorithm by \citet[Eqs.~10\&11]{najfeld95expm_deriv} and the identity
$\frac{\partial\text{ld}_{\MK}(\MW)}{\partial\theta_{j}}=
 \sum_{i}\frac{1}{z_{i}}\frac{\partial z_{i}}{\partial\theta_{j}}.$

Second, the noise precision derivative is computed using the matrix determinant lemma
\[
\frac{\partial\text{ld}_{\MK}(\MW)}{\partial\MW}=\text{diag}(\MG\MR^{-1}\MG^{\top})
\]
where $\MR$ and $\MG$ are as defined in Sec.~\ref{subsub:linsolve}. Since $\MG$ is a Kronecker product, we do not need to know $\MR^{-1}$ completely; only the block diagonal part needs to be evaluated \citep[Sec.~3.1]{grigorievskiy17sparseGPprecision}, which we achieve using the \texttt{sparseinv} package \cite{davis14sparseinv}.

\subsubsection{Predictions}
\label{subsub:pred}
Once the parameters $\valpha$ and $\MW$ have been obtained from one of the inference algorithms, predictions can be computed using Kalman filtering (Alg.~\ref{alg:kalman}) followed by RTS smoothing (Alg.~\ref{alg:rts}) in linear time. The unseen test input(s) $t_*$ are simply included into the data set, then the latent distribution can be extracted via $\sigma^2_{f,i}=\MH \vm_i$ and $\sigma^2_{f,i}=\MH \MQ_i \MH^\T$. Assumed density filtering can be achieved by switching on the ADF flag in Algorithm~\ref{alg:kalman}.

Now that we have detailed the computational primitives, we describe how to use them to drive different approximate inference methods.

\subsection{Laplace approximation (LA)}
\label{sub:la}
The GP Laplace approximation \cite{williams98laplace} is essentially a second order Taylor expansion of the GP posterior $\mathbb{P}(\MF|\mathbf{y})\propto\mathcal{N}(\MF|\mathbf{m},\MK)\prod_{i}\mathbb{P}(y_{i}|f_{i})$ around its mode $\hat{\MF}=\arg\max_{\MF}\mathbb{P}(\MF|\mathbf{y})$ with $W_{ii}=-\partial^{2}\log \mathbb{P}(y_{i}|f_{i})/\partial f_{i}^{2}$ the likelihood curvature and 
\begin{multline*}
  \log Z_{LA}= \\
  -\frac{1}{2}\bigg[\valpha^{\top}\text{mvm}_{\MK}(\valpha)
									+\text{ld}_{\MK}(\MW)
                  -2\sum_{i}\log\mathbb{P}(y_{i}|\hat{f}_{i})\bigg]
\end{multline*}
being an approximation to the (log) marginal likelihood. In practice, we
use a Newton method with line searches. Similar primitives have been
used in Kalman-based demand forecasting \cite{seeger16forecast} with linear models.
Note that for log-concave likelihoods, the mode finding is a convex program.

\subsection{Variational Bayes (VB)}
\label{sub:vb}
The VB method uses convex duality to exactly represent the individual (log) likelihoods as a maximum over quadratics $\ell(f_{i})=\log\mathbb{P}(y_{i}|f_{i})=\max_{W_{ii}}b_{i}f_{i}-W_{ii}f_{i}^{2}/s+h(W_{ii})$ given that the likelihood is super Gaussian (e.g.\ Laplace, Student's $t$, logistic) \cite{gibbs00vb}. Finally, inference can be interpreted as a sequence of Laplace approximations \cite{seeger11scalable} with the smoothed log likelihood  $\ell_\mathrm{VB}(f_{i})=\ell(g_{i})+b_{i}(f_{i}-g_{i})$ with smoothed latent $g_{i}=\mathrm{sign}(f_{i}-z_{i})\sqrt{(f_{i}-z_{i})^{2}+v_{i}}+z_{i}$.
The parameters $(z_{i},b_{i})$ depend on the likelihood only e.g.\ $(z_{i},b_{i})=(y_{i},0)$ for Student's $t$ and Laplace and $(z_{i},b_{i})=(0,y_{i}/2)$ for logistic likelihood and $v_{i}$ is the marginal variance. The marginal likelihood lower bound takes the form 
\begin{multline*}
  \log Z \ge \log Z_\mathrm{VB}= \\
  -\frac{1}{2}\left[\valpha^{\top}\text{mvm}_{\MK}(\valpha)
                 +\text{ld}_{\MK}(\MW)
								 -2\sum_{i}\ell_\mathrm{VB}(f_{i})
								 -2\rho_\mathrm{VB}\right],
\end{multline*}
where $\rho_\mathrm{VB}$ collects a number of scalar terms depending on $(\vz,\vb,\MW,\valpha,\vm)$.

\subsection{Direct Kullback--Leibler minization (KL)}
\label{sub:kl}
Finding the best Gaussian posterior approximation $\mathcal{N}(\vb|\vmu,\MV)$ by minimizing its Kullback--Leibler divergence to the exact posterior is a very generic inference approach \cite{opper09kl} which has recently been made practical via a conjugate variational inference algorithm \cite{khan17cvi} operating as a sequence of GP regression steps. In particular, GP regression problems $j=1,\ldots,J$ are solved for a sequence of Gaussian pseudo observations whose mean and precision $(\tilde{\vy}_{j},\tilde{\MW}_{j})$
are iteratively updated based on the first and second derivative of the convolved likelihood $\ell_\mathrm{KL}(f_{i})=\int\ell(t)\mathcal{N}(f_{i}|t,v_{i})\,\mathrm{d}t$
where $v_{i}$ is the marginal variance until convergence. The marginal
likelihood lower bound takes the form 
\begin{multline*}
  \log Z\ge\log Z_\mathrm{KL} = \\
  -\frac{1}{2}\left[\valpha^{\top}\text{mvm}_{\MK}(\valpha)+\text{ld}_{\MK}(\MW)-2\sum_{i}\ell_\mathrm{KL}(f_{i})-2\rho_\mathrm{KL}\right],
\end{multline*}
where the remainder $\rho_\mathrm{KL} = \mathrm{tr}(\MW^{\top} \partial\text{ld}_{\MK}(\MW)/\partial\MW )$ can be computed using computational primitive 4.

\subsection{Assumed density filtering (ADF) a.k.a.\ single-sweep Expectation propagation (EP)}
\label{sub:adf}
In expectation propagation (EP) \cite{minka01}, the non-Gaussian likelihoods $\mathbb{P}(y_{i}|f_{i})$ are replaced by unnormalized Gaussians $t_{i}(f_{i})=\exp(b_{i}f_{i}-W_{ii}f_{i}^{2}/2)$ and their parameters $(b_{i},W_{ii})$ are iteratively (in multiple passes) updated such that $\mathbb{Q}_{\neg i}(f_{i})\mathbb{P}(y_{i}|f_{i})$ and $\mathbb{Q}_{\neg i}(f_{i})t(f_{i})$ have $k=0,\ldots,2$ identical moments $z_{i}^{k}=\int f_{i}^{k}\mathbb{Q}_{\neg i}(f_{i})t(f_{i})\,\mathrm{d}f_{i}$. Here,  $\mathbb{Q}_{\neg i}(f_{i})=\int\mathcal{N}(\vf|\vm,\MK)\prod_{j\ne i}t_{j}(f_{j})\,\mathrm{d}\vf_{\neg i}$ denotes the cavity distribution. Unlike full state space EP using forward and backward passes \cite{heskes2002expectation}, there is a single-pass variant doing only one forward sweep that is know as assumed density filtering (ADF). It is very simple to implement in the GP setting. In fact, ADF is readily implemented by Algorithm~\ref{alg:kalman} when the flag ``ADF'' is switched on. The marginal likelihood approximation takes the form 
\begin{multline*}
  \log Z_\mathrm{ADF}= \\
-\frac{1}{2}\left[\valpha^{\top}\text{mvm}_{\MK}(\valpha)+\text{ld}_{\MK}(\MW)-2\sum_{i}\log z^0_i-2\rho_\mathrm{ADF}\right],
\end{multline*}
where the remainder $\rho_\mathrm{ADF}$ collects a number of scalar terms depending on $(\vb,\MW,\valpha,\vm)$.

\section{Experiments}
\label{sec:experiments}
The experiments focus on showing that the state space formulation delivers the exactness of the full na\"ive solution, but with appealing computational benefits, and wide applicability over GP regression and classification tasks. Sec.~\ref{sub:interp_exper} assesses the effects of the fast approximations of $\MA_i$ and $\MQ_i$. Sec.~\ref{sub:computations} demonstrates the unprecedented computational speed, and Sec.~\ref{sub:likelihoods} presents a comparison study including 12 likelihood/inference combinations. Finally, two large-scale real-data examples are presented and solved on a standard laptop in a matter of minutes.

\begin{figure}[!t]
  \centering\small
  \pgfplotsset{yticklabel style={rotate=90}, ylabel style={yshift=-15pt}, legend style={font=\scriptsize},grid=both,grid style=dotted}
  \setlength{\figurewidth}{\columnwidth}
  \setlength{\figureheight}{0.75\figurewidth}
  
\begin{tikzpicture}

\begin{axis}[%
xmin=0,
xmax=160,
xlabel={Number of interpolation grid points, $K$},
ymode=log,
ymin=1e-14,
ymax=0.01,
yminorticks=true,
ylabel={Relative absolute difference in $\log Z$},
axis background/.style={fill=white},
legend style={legend cell align=left,align=left,draw=white!15!black},
width=\figurewidth,
height=\figureheight
]
\addplot [color=black,solid,mark size=0.8pt,mark=*,mark options={solid},forget plot]
 plot [error bars/.cd, y dir = both, y explicit]
 table[row sep=crcr, y error plus index=2, y error minus index=3]{%
6	0.000817517098717278	0.00288041246613215	0.000809433638766426\\
10	4.20012209526399e-06	5.5176395097631e-06	3.30859307961377e-06\\
20	7.54265624137765e-08	2.343958216191e-07	6.80181125405908e-08\\
30	1.22308548843621e-08	2.2665114448321e-08	1.22152309765759e-08\\
40	5.67837753562049e-09	1.53214854648987e-08	5.23681872016142e-09\\
50	3.90494436039896e-09	4.62320414204529e-09	3.78684111989787e-09\\
60	1.86091039362687e-09	1.99858363619959e-09	1.56040224634027e-09\\
70	1.24128746539889e-09	2.38895852200951e-09	9.73662663933048e-10\\
80	5.9733839665778e-10	8.77912984985816e-10	5.96826484179785e-10\\
90	5.11908187949143e-10	1.09126212904051e-09	5.06716584346196e-10\\
100	2.0051816867198e-10	6.72934502833922e-10	1.96674348519531e-10\\
110	2.38884766868299e-10	4.11181487234342e-10	2.21788130205568e-10\\
120	2.09094003218628e-10	9.85735090408747e-10	1.60313092403594e-10\\
130	1.39413994836044e-10	2.20454301518357e-10	1.37548994777813e-10\\
140	1.56173996283604e-10	3.48307908142979e-10	1.54716778084177e-10\\
150	9.77907452818168e-11	1.87232825791856e-10	9.52317002491149e-11\\
};
\end{axis}
\end{tikzpicture}%
  
  \vspace*{-1em}
  \caption{Relative differences in $\log Z$ with different approximation grid sizes for $\MA_i$ and $\MQ_i$, $K$, of solving a GP regression problem. Results calculated over 20 independent repetitions, mean$\pm$min/max errors visualized.}
  \label{fig:interpolation}
  \vspace*{-1em}
\end{figure}
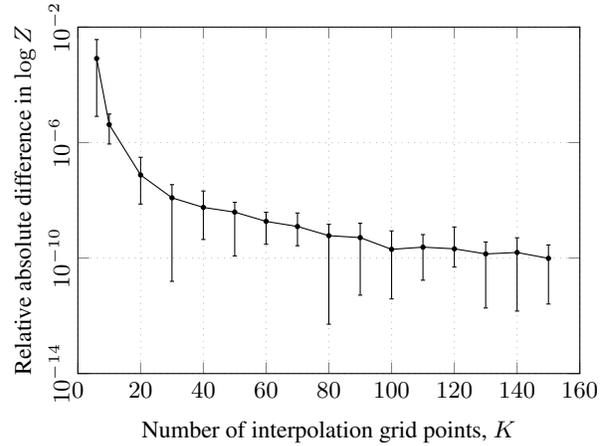

\begin{figure}[!t]
  \centering\small
  \pgfplotsset{yticklabel style={rotate=90}, ylabel style={yshift=-15pt}, legend style={font=\scriptsize},grid=both,grid style=dotted, scaled x ticks=base 10:-3,axis on top}
  \setlength{\figurewidth}{\columnwidth}
  \setlength{\figureheight}{0.75\figurewidth}
  
\begin{tikzpicture}

\begin{axis}[%
unbounded coords=jump,
xmin=100,
xmax=20000,
xlabel={Number of training inputs, $n$},
ymin=0,
ymax=10,
ylabel={Evaluation time (s)},
axis background/.style={fill=white},
legend style={legend cell align=left,align=left,draw=white!15!black},
width=\figurewidth,
height=\figureheight
]
\addplot [color=white!60!black,solid]
 plot [error bars/.cd, y dir = both, y explicit]
 table[row sep=crcr, y error plus index=2, y error minus index=3]{%
500	0.0542559705	0.1071538315	0.0040126145\\
750	0.0713655395	0.0262773375	0.0023100315\\
1000	0.131048806	0.007239309	0.00319993199999999\\
1500	0.3043723545	0.00581775950000002	0.00358580250000001\\
2000	0.575775973	0.0260291739999999	0.00957825400000001\\
2500	0.9728757535	0.0421437535000001	0.00907602949999997\\
3000	1.5065125805	0.0468537785000001	0.00422406649999996\\
3500	2.2059819015	0.6211062935	0.0175419204999998\\
4000	3.0795903285	0.0376089815	0.0181349365000001\\
4500	4.5305254085	0.0637963435	0.0139158894999998\\
5000	5.930630445	0.0223168849999995	0.0548990659999999\\
6000	9.337087073	0.0902762790000011	0.0307328529999999\\
7000	nan	nan	nan\\
8000	nan	nan	nan\\
9000	nan	nan	nan\\
10000	nan	nan	nan\\
12000	nan	nan	nan\\
14000	nan	nan	nan\\
16000	nan	nan	nan\\
18000	nan	nan	nan\\
20000	nan	nan	nan\\
};
\addlegendentry{Na\"ive};

\addplot [color=black,solid]
 plot [error bars/.cd, y dir = both, y explicit]
 table[row sep=crcr, y error plus index=2, y error minus index=3]{%
500	0.158573289	0.160156516	0.006910686\\
750	0.234131097	0.025009727	0.016270124\\
1000	0.2977783535	0.0118414775	0.00713750450000006\\
1500	0.4491071815	0.0163096835	0.00812243350000003\\
2000	0.5970470235	0.0385055715	0.00891671050000009\\
2500	0.7553048225	0.0166602934999999	0.00974638250000004\\
3000	0.8948417215	0.0242583645000001	0.00354463350000001\\
3500	1.044985192	0.0444904910000001	0.021609258\\
4000	1.215330433	0.0138041359999999	0.0220601450000002\\
4500	1.296618213	0.0193911490000001	0.0304872279999999\\
5000	1.4467958845	0.0185011004999998	0.0458911515000002\\
6000	1.744892205	0.01771689	0.0525066839999999\\
7000	1.994787335	0.0643903989999999	0.0131523730000001\\
8000	2.292190978	0.0540151790000003	0.031444107\\
9000	2.5975550875	0.0518432775000002	0.0278536374999998\\
10000	2.8690862395	0.0346077624999999	0.0361462035\\
12000	3.445348023	0.042097627	0.0384506720000002\\
14000	4.002665895	0.0658266550000004	0.0356733869999997\\
16000	4.619203373	0.0339186050000002	0.0808477069999993\\
18000	5.1622150825	0.0850491225000001	0.0320757374999996\\
20000	5.7368808035	0.0391088574999996	0.0669853475000002\\
};
\addlegendentry{State space};

\addplot [color=black,dashed]
 plot [error bars/.cd, y dir = both, y explicit]
 table[row sep=crcr, y error plus index=2, y error minus index=3]{%
500	0.1558943675	0.0500944255	0.00539759250000002\\
750	0.2290385395	0.0396342505	0.0107943565\\
1000	0.3000520195	0.0136499995	0.00911341650000003\\
1500	0.4469249545	0.0110627914999999	0.0161174555\\
2000	0.597140399	0.016193793	0.0134367569999999\\
2500	0.6596313915	0.0158640354999999	0.0137948835\\
3000	0.7197739645	0.0133515175	0.0216626025000001\\
3500	0.7893052735	0.0402756215000001	0.0161633334999999\\
4000	0.8450944965	0.0238471195	0.0132277015\\
4500	0.9145882935	0.0149394525	0.0183738175\\
5000	0.983402699	0.0198486950000001	0.012303853\\
6000	1.111269694	0.144153731	0.0192535359999999\\
7000	1.241247293	0.0212389470000003	0.0173959279999998\\
8000	1.3645325405	0.0628647275000001	0.0301765144999999\\
9000	1.5156920865	0.0377943454999998	0.0158656825000001\\
10000	1.639859004	0.040378885	0.0116140810000001\\
12000	1.902613168	0.023446831	0.00562805700000002\\
14000	2.1759867355	0.0179924625000001	0.0347925514999998\\
16000	2.4382727505	0.0481512915	0.0373211935\\
18000	2.7212710675	0.0299056604999999	0.0518265714999999\\
20000	2.9857448135	0.0966323714999997	0.0521880855000001\\
};
\addlegendentry{State space ($K=2000$)};

\addplot [color=black,dashdotted]
 plot [error bars/.cd, y dir = both, y explicit]
 table[row sep=crcr, y error plus index=2, y error minus index=3]{%
500	0.079824382	0.069683611	0.002641902\\
750	0.116797673	0.00744217500000001	0.00331510599999998\\
1000	0.1495631235	0.0057653865	0.00744274250000002\\
1500	0.211457341	0.016764303	0.00466242400000003\\
2000	0.2805918855	0.0192378365	0.0103122275\\
2500	0.340628866	0.00695894100000005	0.00567991499999998\\
3000	0.410200589	0.013334632	0.00778509999999999\\
3500	0.4743740575	0.0345038225	0.0149995315\\
4000	0.537102815	0.013704388	0.010555254\\
4500	0.591848168	0.0280928679999999	0.00816242600000006\\
5000	0.673001681	0.032165182	0.011437936\\
6000	0.800696487	0.00811721200000004	0.013805428\\
7000	0.9280771245	0.0263673525	0.0115807805\\
8000	1.0643306645	0.0246636525000001	0.0279314815\\
9000	1.195344116	0.019776155	0.023013344\\
10000	1.3378384785	0.0186223674999999	0.0449545425000002\\
12000	1.601774833	0.0328856580000001	0.0523004269999998\\
14000	1.856580204	0.024651048	0.0424966270000002\\
16000	2.123899687	0.02502476	0.0234972200000003\\
18000	2.407280593	0.0391334459999997	0.0472454380000005\\
20000	2.654603122	0.0598432889999998	0.029269717\\
};
\addlegendentry{State space ($K=10$)};

\end{axis}
\end{tikzpicture}%
  
  \vspace*{-1em}
  \caption{Empirical computational times of GP prediction using the GPML toolbox implementation as a function of number of training inputs, $n$, and degree of approximation, $K$. For all four methods the maximum absolute error in predicted means was $10^{-9}$. Results calculated over ten independent runs.}
  \label{fig:computations}
  \vspace*{-1em}
\end{figure}
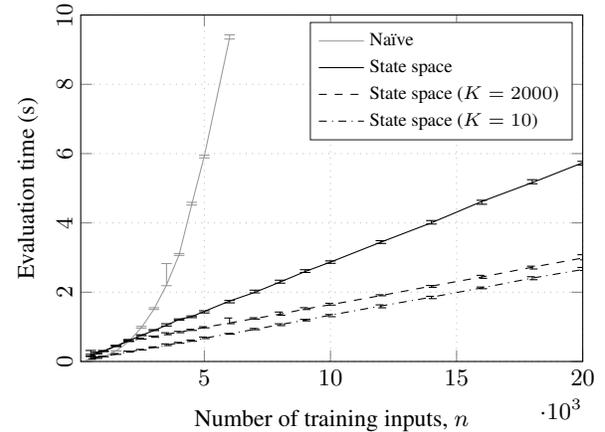

\begin{table*}
\centering\small
\caption{A representative subset of supported likelihoods and inference schemes \citep[for a full list, see][]{rasmussen10gpml}. Results for simulated data with $n=1000$ (around the break-even point of computational benefits). Results compared to respective na\"ive solution in mean absolute error (MAE). $^\dagger$The results for EP are compared against ADF explaining the deviation and speed-up.}
\label{tbl:likinf}
\begin{tabular*}{\textwidth}{@{\extracolsep{\fill}} lllllcccl}
\toprule 
Likelihood & Inference & MAE in $\valpha$ & MAE in $\MW$ & MAE in $\vmu_{f,*}$ & $-\log Z$ & $-\log Z_\mathrm{ss}$ & $t/t_\mathrm{ss}$ & Description \\
\midrule
Gaussian    & Exact        & $<10^{-4}$ & $<10^{-16}$ & $<10^{-14}$ & $-1252.29$ & $-1252.30$ & $2.0  $ & Regression \\
Student's $t$ & Laplace      & $<10^{-7}$ & $<10^{-6}$ & $<10^{-3}$ & $2114.45$ & $2114.45$ & $1.4  $ & Regression, \\
Student's $t$ & VB           & $<10^{-6}$ & $<10^{-6}$ & $<10^{-7}$ & $2114.72$ & $2114.72$ & $2.7  $ & \hspace{1em}robust \\
Student's $t$ & KL           & $<10^{-4}$ & $<10^{-4}$ & $<10^{-5}$ & $2114.86$ & $2114.86$ & $4.6  $ &  \\
Poisson     & Laplace      & $<10^{-6}$ & $<10^{-4}$ & $<10^{-6}$ & $1200.11$ & $1200.11$ & $1.2  $ & Poisson regression, \\
Poisson     & EP/ADF$^\dagger$ & $<10^{-1}$ & $<10^{0}$ & $<10^{-2}$ & $1200.11$ & $1206.59$ & $39.5 $ & \hspace{1em}count data \\
Logistic    & Laplace      & $<10^{-8}$ & $<10^{-7}$ & $<10^{-7}$ & $491.58$ & $491.58$ & $1.3  $ & Classification, \\
Logistic    & VB           & $<10^{-6}$ & $<10^{-6}$ & $<10^{-6}$ & $492.36$ & $492.36$ & $2.3  $ & \hspace{1em}logit regression \\
Logistic    & KL           & $<10^{-7}$ & $<10^{-6}$ & $<10^{-7}$ & $491.57$ & $491.57$ & $4.0  $ &  \\
Logistic    & EP/ADF$^\dagger$ & $<10^{-1}$ & $<10^{0}$ & $<10^{-1}$ & $491.50$ & $525.46$ & $48.1 $ &  \\
Erf         & Laplace      & $<10^{-8}$ & $<10^{-6}$ & $<10^{-7}$ & $392.01$ & $392.01$ & $1.2  $ & Classification, \\
Erf         & EP/ADF$^\dagger$ & $<10^{0}$ & $<10^{0}$ & $<10^{-1}$ & $392.01$ & $433.75$ & $37.1 $ & \hspace{1em}probit regression \\
\bottomrule
\end{tabular*}
\vspace{-1em}
\end{table*}

\subsection{Effects in fast computation of $\MA_i$ and $\MQ_i$}
\label{sub:interp_exper}
In the first experiment we study the validity of the interpolation to approximate matrix exponential computation (Sec.~\ref{sub:fast_interp}). The input time points of observations $t_i$ were randomly selected from the interval [0, 12] and outputs $y_i$ were generated from the sum of two sinusoids plus Gaussian noise: $y_i = 0.2 \sin(2\pi \, t_i + 2) + 0.5 \sin( 0.6 \pi \, t_i + 0.13) + 0.1 \, \mathcal{N}(0,1)$. The $\Delta t_i$s were exponentially distributed since the time points followed a Poisson point process generation scheme. All results were calculated over 20 independent realizations.

For each generated dataset we considered GP regression (in the form of Sec.~\ref{sub:ss_primitives}) with a Gaussian likelihood and Mat\'ern ($\nu=5/2$) covariance function. Initially, all the matrices $\MA_i$ and $\MQ_i$ were computed exactly. The results were compared to the approximate results of those matrices with various number of interpolation grid points $K$. The absolute relative difference between the approximated and not approximated marginal likelihood and its derivatives were computed. The results are given in Figure~\ref{fig:interpolation}. The figure shows that the relative difference is decreasing with the number of grid points and finally saturates. Hence, increasing accuracy of approximation with the growing size of the interpolation grid. More figures with the accuracies of the derivatives computations can be found in the Supplementary material.

\subsection{Computational benefits}
\label{sub:computations}
The practical computational benefits of the state space form in handling the latent were evaluated in the following simulation study. We consider GP regression with a Mat\'ern ($\nu=3/2$) covariance function with simulated data from a modified sinc function ($6\sin(7\pi\,x)/(7\pi\,x+1)$) with Gaussian measurement noise and input locations $x$ drawn uniformly. The number of data points was increased step-wise from $n=500$ to $n=\text{20,000}$. The calculations were repeated for 10 independent realizations of noise.

The results (including results in following sections) were run on an Apple MacBook Pro (2.3~GHz Intel Core i5, 16~Gb RAM) laptop in Mathworks Matlab 2017b. All methods were implemented in the GPML Toolbox framework, and the state space methods only differed in terms of solving the continuous-time model for $\MA_i$ and $\MQ_i$ (see Sec.~\ref{sub:fast_interp}). 

Figure~\ref{fig:computations} shows the empirical computation times for the $\mathcal{O}(n^3)$ na\"ive and $\mathcal{O}(n)$ state space results. The state space results were computed with no interpolation, and $\MA_i$ and $\MQ_i$ interpolated with $K=2000$ and $K=10$. The computation times with $K=2000$ follow the exact state space model up to $n=2000$. In terms of error in predictive mean over a uniform grid of 200 points, the maximum absolute error of state space results compared to the na\"ive results was $10^{-9}$.

\subsection{Numerical effects in non-Gaussian likelihoods}
\label{sub:likelihoods}
The previous section focused on showing that the latent state space computations essentially exact up to numerical errors or choices of interpolation factors in solving the continuous-time model. Delivering the computational primitives for approximate inference using LA, VB, KL, or EP should thus give the same results as if run through na\"ively.

Table~\ref{tbl:likinf} shows a representative subset of combinations of likelihoods and inference scheme combinations \citep[for a full list, see][]{rasmussen10gpml}. For each model, appropriate data was produced by modifying the simulation scheme explained in the previous section (Student's $t$: 10\% of observations outliers; Poisson: counts followed the exponentiated sinc function; Logistic/Erf: the sign function applied to the sinc). The mean absolute error in $\valpha$, $\MW$, and $\vmu_{f,*}$ between the state space and na\"ive solution are shown. The results are equal typically up to 4--6 decimals. It is probable that the state space approach shows accumulation of numerical errors. The large offsets in the EP values are due to our state space implementation being single-sweep (ADF). Here only $n=1000$ data points were used, while Figure~\ref{fig:computations} shows that for regression the computational benefits only really kick-in in around $n=2000$. For example in KL, the speed-up is clear already at $n=1000$.

\subsection{Robust regression of electricity consumption}
\label{sub:electricity}
We present a proof-of-concept large-scale robust regression study using a Student's $t$ likelihood for the observations, where the data is inherently noisy and corrupted by outlying observations. We consider hourly observations of log electricity consumption \cite{hebrail12electricity} for one household (in log~kW) over a time-period of 1,442 days ($n = \text{34,154}$, with 434 missing observations). We use a GP with a Student's $t$ likelihood (with one degree of freedom) and a Mat\'ern ($\nu=3/2$) covariance function for predicting/interpolating values for missing days (state dimensionality $d=2$). For inference we use direct KL minimization (Sec.~\ref{sub:kl}). We evaluate our approach by 10-fold cross-validation over complete days, in this experiment with fixed hyperparameters, and obtain a predictive RMSE of $0.98\pm0.02$ and NLPD of $1.47\pm0.01$.

\subsection{Airline accidents}
\label{sub:airline}
Finally, we study the regression problem of explaining the time-dependent intensity of accidents and incidents of commercial aircraft. The data consists of dates of incidents that were scraped form \cite{wikiair18}, and it covers 1210 incidents over the time-span of 1919--2017. We use a log Gaussian Cox process, an inhomogeneous Poisson process model for count data. The unknown intensity function $\lambda(t)$ is modeled with a log-Gaussian process such that $f(t) = \log \lambda(t)$. The likelihood of the unknown function corresponds to
$ \mathbb{P}(\{t_i\} | f) = \exp\left(-\int \exp(f(t))\,\mathrm{d}t + \sum_{i=1}^n f(t_i) \right)$.
However, this likelihood requires non-trivial integration over the exponentiated GP. \citet{moller98logCox} propose a locally constant intensity in subregions based on discretising the interval into bins. This approximation corresponds to having a Poisson model for each bin. The likelihood becomes
$ \mathbb{P}(\{t_i\} | f) \approx \prod_{j=1}^N \mathrm{Poisson}(y_j | \exp(f(\hat{t}_j)))$,
where $\hat{t}_j$ is the bin coordinate and $y_j$ the number of data points in it. This model reaches posterior consistency in the limit of bin width going to zero ($N \to \infty$) \citep{tokdar07postConsist}. Thus it is expected that the results improve the tighter the binning is.

We use a bin width of one day leading to $N = \text{35,959}$ observations, and a prior covariance structure
$ k(t,t') = k_\text{Mat\'ern}(t,t') + k_\text{periodic}(t,t')\,k_\text{Mat\'ern}(t,t') $
capturing a slow trend and decaying time-of-year effect (period one year). The model state dimension is $d=30$. For inference we used ADF (single-sweep EP, Sec.~\ref{sub:adf}). All hyperparameters (except the period length) were optimized w.r.t.\ marginal likelihood, such that we first obtained a ball-park estimate of the parameters using one-month binning, and then continued optimizing with the full data set.

\begin{figure}[!t]
  \centering\small
  \pgfplotsset{yticklabel style={rotate=90}, ylabel style={yshift=-15pt},clip=false,xlabel style={yshift=-6pt},scale only axis}
  \setlength{\figurewidth}{.8\columnwidth}
  \setlength{\figureheight}{.55\columnwidth}
  
  \input{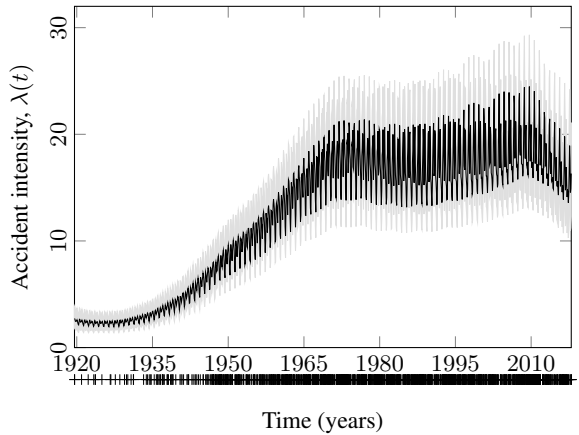}
  
  \vspace*{-1em}
  \caption{Intensity of aircraft incident modeled by a log Gaussian Cox process with the mean and approximate 90\% confidence regions visualized ($N=\text{35,959}$). The observed indicent dates are shown by the markers on the bottom.}
  \label{fig:aircraft}
  \vspace{-.5em}
\end{figure}

\begin{figure}[!t]
  \centering\small
  \pgfplotsset{yticklabel style={rotate=90}, ylabel style={yshift=-15pt},clip=true,scale only axis}
  \setlength{\figurewidth}{.75\columnwidth}
  \setlength{\figureheight}{.55\columnwidth}
  
  \input{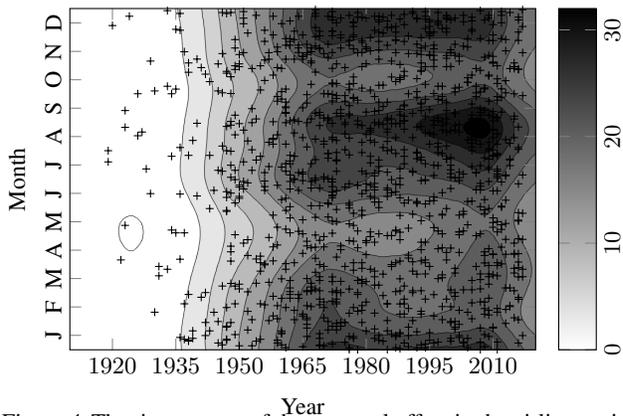}
  
  \vspace*{-2em}
  \caption{The time course of the seasonal effect in the airline accident intensity, plotted in a year vs.\ month plot (with wrap-around continuity between edges). Markers show  incident dates. The bimodal yearly effect has started receding in the previous years.}
  \label{fig:aircraft_seasonal}
  \vspace*{-1em}
\end{figure}

Figure~\ref{fig:aircraft} shows the time-dependent intensity $\lambda(t)$ that show a clear trend and pronounced periodic effects. The time course of the periodic effects are better visible in Figure~\ref{fig:aircraft_seasonal} that show the gradual formation of the periodicity, and the more recent decay of the winter mode. We omit speculation of explaining factors in the data, but assume the effects to be largely explained by the number of operating flights. We further note that a wider bin size would deteriorate the analysis of the periodic peaks (they become `smoothed' out), thus justifying the need for the large $N$ as speculated above.

\section{Discussion and conclusion}
\label{sec:discussion}
Motivated by the computational constraints imposed by analytic intractability in the non-conjugate setting and cubic scaling, we propose to extend the state space representation of Gaussian processes to the non-Gaussian setting. We cast a range of approximate inference schemes using a small set of generic computational primitives to enable a unified treatment and show how to implement them using scalable algorithms relying on Kalman filters and dynamical system theory. We propose to use convolution interpolation to accelerate the expensive matrix exponential computations, which further reduces the runtime by a factor of two. We demonstrate computational benefits on a number of time series datasets to illustrate the tradeoffs and the achievable accuracy as compared to the dense setting. 

Possible drawbacks are related to the cubic computational complexity in model state dimension (e.g.\ when considering several products of covariance functions), and problems related to floating point precision accumulating in the recursions when $n$ is very large.

Overall, we conclude that for accurate scalable inference in GP time series, the state space viewpoint adds a valuable alternative to the computational toolbox of the modeling practitioner using our reference implementation.

\bibliographystyle{icml2018}
\bibliography{related}

\clearpage
\thispagestyle{empty}
\twocolumn[{\center\baselineskip 18pt
   \toptitlebar{\Large\bf {Supplementary Material for} \\ State Space Gaussian Processes with Non-Gaussian Likelihood}\bottomtitlebar} In this appendix we provide further identities that are made possible by the recursive formulation together with some additional plots addressing the effects of possible approximations. The results in A.\ follow as a by-product of the algorithms presented in the main paper, and are provided here as additional material. \vspace*{3em}]

\appendix

\section{Recursions for $\valpha$ and $\ML$}
The lower-triangular Cholesky factor $\ML \in \mathbb{R}^{n \times n}$ given by
\begin{equation}
  \ML \, \ML^\T = \MK + \MW^{-1}
\end{equation}
can in the general case be solved efficiently in $\mathcal{O}(n^3)$. If the covariance function is Markovian, the following recursion can be used for forming the Cholesky factor in $\mathcal{O}(n^2)$ time complexity:
\begin{equation}
  \ML_{i,i} = \sqrt{s_i}
\end{equation}
with $s_i=z_i/W_{ii}$ the innovation variance of Algorithm~\ref{alg:kalman} for the diagonal and 
\begin{equation}
  \ML_{i,j} = \MH \, \bigg[ \prod_{k=i}^{j-1} \MA_{k} \bigg] \, \vk_j \, \sqrt{s_i}
\end{equation}
for the lower-triangular off-diagonal elements, $i=1,2,\ldots,n$ and $j<i$. The matrix product is constructed by iterated right-side multiplication.

The matrix-inverse of the Cholesky factor is also interesting as it gives the inverse of the original expression:
\begin{equation}
  \ML^{-1} \, \ML^{-\mathsf{T}} = (\MK + \MW^{-1})^{-1}.
\end{equation}
The inverse Cholesky factor can be constructed as follows in $\mathcal{O}(n^2)$ time complexity:
\begin{equation}
  [\ML^{-1}]_{i,i} = 1/\sqrt{s_i}
\end{equation}
for the diagonal and 
\begin{equation}
  [\ML^{-1}]_{j,i} = -\MH \, \bigg[ \prod_{k=i}^{j-1} (\MI - \vk_k \, \MH) \bigg] \, \vk_j / \sqrt{s_i}
\end{equation}
for the lower-triangular off-diagonal elements, $i=1,2,\ldots,n$ and $j<i$.

Rather than directly solving $\vbeta = \ML \backslash \vr$ or $\valpha = \ML^\T \backslash (\ML \backslash \vr)$ by solving the linear systems by forward and backward substitution (in $\mathcal{O}(n^2)$) using the Cholesky factor $\ML$ obtained in the previous section, the vectors $\valpha$ and $\vbeta$ can be formed in $\mathcal{O}(n)$ time complexity (and $\mathcal{O}(n)$ memory) by the following forward and backward recursions using the filter forward and smoother backward passes.

The recursion for forward solving $\vbeta \in \mathbb{R}^n$:
\begin{equation}
  \beta_i = v_i / \sqrt{s_i},
\end{equation}
where $v_i=-c_i/W_{ii}$ and $s_i=z_i/W_{ii}$ are the Kalman filter (Alg.~\ref{alg:kalman}) innovation mean and variance at step $i=1,2,\ldots,n$.

The calculation of $\valpha$ can most easily be done by utilizing the Rauch--Tung--Striebel mean and gain terms as follows:
\begin{equation}
  \alpha_i = \beta_i/\sqrt{s_i} - W_{ii}\MH \Delta\vm_{i} = \gamma_i - W_{ii}\MH \Delta\vm_{i},
\end{equation}
for $i = 1,2,\ldots, n-1$ and $\alpha_n = \beta_n/\sqrt{s_n} = \gamma_n$. 

\section{Extra results of experiments in Sec.~\ref{sub:interp_exper}}
Figure~\ref{fig:interpolation_extra} provides additional plots for the interpolation experiment study in the main paper. The effects induced by approximations in solving $\MA_i$ and $\MQ_i$ are more pronounced for small $K$, when comparing the derivative terms (w.r.t.\ hyperparameters) of $\log Z$. Even for the derivative terms the errors drop quickly as a function of approximation degree $K$.

\begin{figure*}[!t]
  \centering\small
  \pgfplotsset{yticklabel style={rotate=90}, ylabel style={yshift=-15pt}, legend style={font=\scriptsize},grid=both,grid style=dotted}
  \setlength{\figurewidth}{.32\textwidth}
  \setlength{\figureheight}{\figurewidth}
  
  \begin{subfigure}[b]{.3\textwidth}
\begin{tikzpicture}

\begin{axis}[%
xmin=0,
xmax=160,
xlabel={Number of interpolation grid points, $K$},
ymode=log,
ymin=1e-08,
ymax=100000000,
yminorticks=true,
ylabel={Relative absolute difference},
axis background/.style={fill=white},
legend style={legend cell align=left,align=left,draw=white!15!black},
width=\figurewidth,
height=\figureheight
]
\addplot [color=black,solid,forget plot]
 plot [error bars/.cd, y dir = both, y explicit]
 table[row sep=crcr, y error plus index=2, y error minus index=3]{%
6	89803.6596003257	1206508.94762113	88500.5166709982\\
10	112.160727726906	1628.18242440109	111.821467568031\\
20	2.40536659655311	32.1821974869123	2.40203709263298\\
30	0.440924593868006	5.82323839641972	0.440550202893471\\
40	0.156416908292007	1.70120337394671	0.154554222992568\\
50	0.0973827583139964	1.28539972505899	0.0969052907631128\\
60	0.0355885776169883	0.375638738943601	0.035443455848724\\
70	0.0178218059969963	0.250257303663738	0.0177691829136946\\
80	0.00370256580898178	0.0104574072700737	0.00365944918558415\\
90	0.0159882253493218	0.253787296880896	0.01598628093022\\
100	0.0104213826935665	0.160900530526022	0.0103523216023758\\
110	0.00816108363949362	0.121921362800308	0.00814063933533711\\
120	0.00319662529447197	0.0292161556756704	0.00319317704195131\\
130	0.00343323852138916	0.0387709230446162	0.00343313889566877\\
140	0.00471965617317434	0.0652157278871386	0.0047148929038634\\
150	0.00116425798346371	0.0136925505075641	0.00116252261160048\\
};
\end{axis}
\end{tikzpicture}%
    \caption{Derivative w.r.t.\ $\ell$}
  \end{subfigure}  
  \hspace*{\fill}  
  \begin{subfigure}[b]{.3\textwidth}
\begin{tikzpicture}

\begin{axis}[%
xmin=0,
xmax=160,
xlabel={Number of interpolation grid points, $K$},
ymode=log,
ymin=1e-08,
ymax=1000000,
yminorticks=true,
ylabel={Relative absolute difference},
axis background/.style={fill=white},
legend style={legend cell align=left,align=left,draw=white!15!black},
width=\figurewidth,
height=\figureheight
]
\addplot [color=black,solid,forget plot]
 plot [error bars/.cd, y dir = both, y explicit]
 table[row sep=crcr, y error plus index=2, y error minus index=3]{%
6	15492.4478942282	197580.01040934	14792.0796134394\\
10	13.7103609680104	171.422907266964	13.4668373204744\\
20	0.197316597265628	1.61365562541213	0.196981940484774\\
30	0.0673158633257704	0.905698482389926	0.0668461173182238\\
40	0.0305049728474501	0.434365437380298	0.030279890508684\\
50	0.0192370737393758	0.267976092500576	0.0190780254635303\\
60	0.00593762526104663	0.0719808110655697	0.00582974430623258\\
70	0.00255999223579072	0.0238659644828797	0.00253430877844289\\
80	0.00329354025312906	0.0485200962572905	0.00329342476902935\\
90	0.00097376939505552	0.0104224686485318	0.000966724163494173\\
100	0.000664997426006832	0.00282283756968328	0.000639330300862592\\
110	0.000540404712099278	0.00399656543405576	0.000533086563858588\\
120	0.000469505473385102	0.00318539571948025	0.000458745793852526\\
130	0.000200468981546256	0.00148108318323299	0.00019951950867698\\
140	0.00113193548149747	0.0156012078136814	0.0011234777532574\\
150	0.000679703399307164	0.00929314278862213	0.000676006576335589\\
};
\end{axis}
\end{tikzpicture}%
    \caption{Derivative w.r.t.\ $\sigma_f$}    
  \end{subfigure}  
  \hspace*{\fill}
  \begin{subfigure}[b]{.3\textwidth}
\begin{tikzpicture}

\begin{axis}[%
xmin=0,
xmax=150,
xlabel={Number of interpolation grid points, $K$},
ymode=log,
ymin=1e-06,
ymax=1000000,
yminorticks=true,
ylabel={Relative absolute difference},
axis background/.style={fill=white},
legend style={legend cell align=left,align=left,draw=white!15!black},
width=\figurewidth,
height=\figureheight
]
\addplot [color=black,solid,forget plot]
 plot [error bars/.cd, y dir = both, y explicit]
 table[row sep=crcr, y error plus index=2, y error minus index=3]{%
6	9234.76335064817	44206.8145083356	8750.4920234415\\
10	11.7537986824134	55.7776004578702	11.4780461255771\\
20	0.189897693183803	1.08576974472353	0.17736727556471\\
30	0.0629604845863586	0.334713903869071	0.0618309937629689\\
40	0.0259891751144473	0.191470596215193	0.025505445377391\\
50	0.0155842101514732	0.0911354815546875	0.0155344839397987\\
60	0.00612857431065486	0.0258465595917882	0.00606148070798249\\
70	0.00313529699427701	0.0123133389403808	0.0030309668285666\\
80	0.00190803999580845	0.0052645703915188	0.0019037349381048\\
90	0.00131308926524932	0.00866936072443972	0.00130718968289381\\
100	0.000526206372429886	0.00367214892036684	0.000524677955455576\\
110	0.000725294729445866	0.00377731309038919	0.000711739633635841\\
120	0.00031491282542034	0.000698372093873349	0.000305991693270973\\
130	0.000363702943617716	0.00136925660607542	0.000362068385948348\\
140	0.000436796344430185	0.00188692392595124	0.000430981633515139\\
150	0.000376715685734588	0.00169422750976175	0.00036690050861073\\
};
\end{axis}
\end{tikzpicture}%
    \caption{Noise scale derivative}
  \end{subfigure}      
  
  \vspace*{-1em}
  \caption{Relative absolute differences in derivatives of $\log Z$ with respect to covariance hyperparameters and noise scale. Different approximation grid sizes, $K$, for solving $\MA_i$ and $\MQ_i$ regression are evaluated. Results calculated over 20 independent repetitions, mean$\pm$min/max errors visualized.}
  \label{fig:interpolation_extra}
\end{figure*}

\end{document}